\documentclass[final]{cvpr}

\usepackage{times}
\usepackage{epsfig}
\usepackage{graphicx}
\usepackage{amsmath}
\usepackage{amssymb}

\usepackage{subfigure}
\usepackage{multirow}
\usepackage[pagebackref=true,breaklinks=true,colorlinks,bookmarks=false]{hyperref}

\def\cvprPaperID{5445} 
\def\confYear{CVPR 2021}

\begin{document}

\title{Semantic Relation Reasoning for Shot-Stable Few-Shot Object Detection}

\author{Chenchen Zhu \quad Fangyi Chen \quad Uzair Ahmed \quad Zhiqiang Shen \quad Marios Savvides\\
Carnegie Mellon University\\
{\tt\small \{chenchez, fangyic, uzaira, zhiqians, marioss\}@andrew.cmu.edu}
}

\maketitle

\begin{abstract}
Few-shot object detection is an imperative and long-lasting problem due to the inherent long-tail distribution of real-world data. Its performance is largely affected by the data scarcity of novel classes. But the semantic relation between the novel classes and the base classes is constant regardless of the data availability. In this work, we investigate utilizing this semantic relation together with the visual information and introduce explicit relation reasoning into the learning of novel object detection. Specifically, we represent each class concept by a semantic embedding learned from a large corpus of text. The detector is trained to project the image representations of objects into this embedding space. We also identify the problems of trivially using the raw embeddings with a heuristic knowledge graph and propose to augment the embeddings with a dynamic relation graph. As a result, our few-shot detector, termed SRR-FSD, is robust and stable to the variation of shots of novel objects. Experiments show that SRR-FSD can achieve competitive results at higher shots, and more importantly, a significantly better performance given both lower explicit and implicit shots. The benchmark protocol with implicit shots removed from the pretrained classification dataset can serve as a more realistic setting for future research. 
\end{abstract}

\section{Introduction}
Deep learning algorithms usually require a large amount of annotated data to achieve superior performance. To acquire enough annotated data, one common way is by collecting abundant samples from the real world and paying annotators to generate ground-truth labels. However, even if all the data samples are well annotated based on our requirements, we still face the problem of few-shot learning. Because long-tail distribution is an inherent characteristic of the real world, there always exist some rare cases that have just a few samples available, such as rare animals, uncommon road conditions. In other words, we are unable to alleviate the situation of scarce cases by simply spending more money on annotation even big data is accessible. Therefore, the study of few-shot learning is an imperative and long-lasting task.

\begin{figure}
    \centering
    \includegraphics[width=\columnwidth]{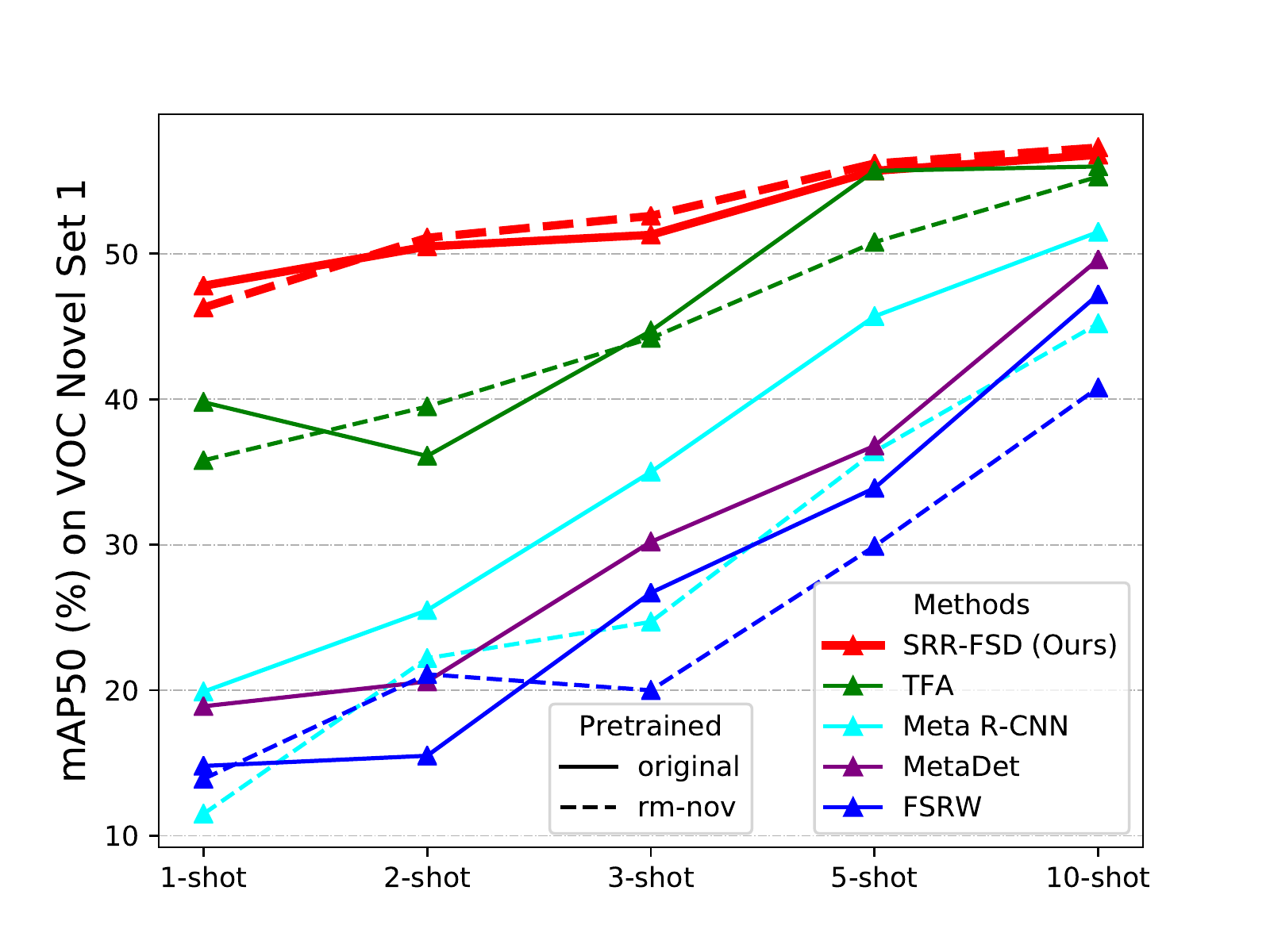}
    \caption{FSOD performance (mAP50) on VOC \cite{voc} Novel Set 1 at different shot numbers. Solid line (original) means the pretrained model used for initializing the detector backbone is trained on the original ImageNet \cite{imagenet}. Dashed line (rm-nov) means classes in Novel Set 1 are removed from the ImageNet for the pretrained backbone model. Our SRR-FSD is more stable to the variation of explicit shots (x-axis) and implicit shots (original vs. rm-nov).}
    \label{fig:vocnov1}
\end{figure}

Recently, efforts have been put into the study of few-shot object detection (FSOD) \cite{lstd, repmet, fsod-mc, yolo-fewshot, meta-rcnn, metadet, attention-rpn, context-transformer, tfa, mpsr, fsdetview}. In FSOD, there are base classes in which sufficient objects are annotated with bounding boxes and novel classes in which very few labeled objects are available. The novel class set does not share common classes with the base class set. The few-shot detectors are expected to learn from limited data in novel classes with the aid of abundant data in base classes and to be able to detect all novel objects in a held-out testing set. To achieve this, most recent few-shot detection methods adopt the ideas from meta-learning and metric learning for few-shot recognition and apply them to conventional detection frameworks, e.g. Faster R-CNN \cite{faster-rcnn}, YOLO \cite{yolov2}. 


Although recent FSOD methods have improved the baseline considerably, data scarcity is still a bottleneck that hurts the detector's generalization from a few samples. In other words, the performance is very sensitive to the number of both explicit and implicit shots and drops drastically as data becomes limited. The explicit shots refer to the available labeled objects from the novel classes. For example, the 1-shot performance of some FSOD methods is less than half of the 5-shot or 10-shot performance, as shown in Figure \ref{fig:vocnov1}. In terms of implicit shots, initializing the backbone network with a model pretrained on a large-scale image classification dataset is a common practice for training an object detector. However, the classification dataset contains many implicit shots of object classes overlapped with the novel classes. So the detector can have early access to novel classes and encode their knowledge in the parameters of the backbone. Removing those implicit shots from the pretrained dataset also has a negative impact on the performance as shown in Figure \ref{fig:vocnov1}. The variation of explicit and implicit shots could potentially lead to system failure when dealing with extreme cases in the real world.

\begin{figure}
    \centering
    \includegraphics[width=\columnwidth]{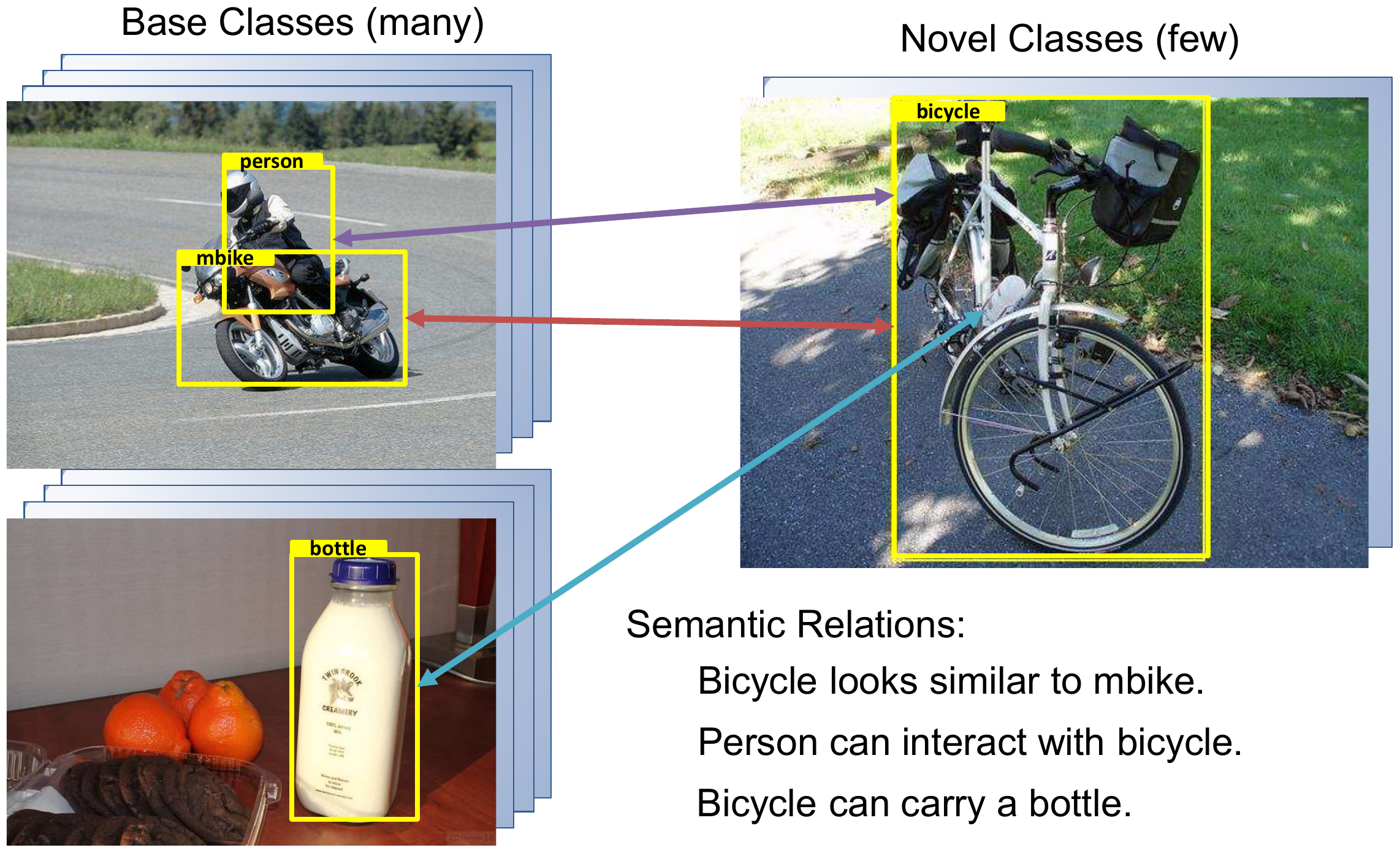}
    \caption{Key insight: the semantic relation between base and novel classes is constant regardless of the data availability of novel classes, which can aid the learning together with visual information.}
    \label{fig:semantic_relation}
\end{figure}

We believe the reason for shot sensitivity is due to exclusive dependence on the visual information. Novel objects are learned through images only and the learning is independent between classes. As a result, visual information becomes limited as image data becomes scarce. However, one thing remains constant regardless of the availability of visual information, i.e. the semantic relation between base and novel classes. For example in Figure \ref{fig:semantic_relation}, if we have the prior knowledge that the novel class ``bicycle'' looks similar to ``motorbike'', can have interaction with ``person'', and can carry a ``bottle'', it would be easier to learn the concept ``bicycle'' than solely using a few images. Such explicit relation reasoning is even more crucial when visual information is hard to access \cite{zsr-gnn}.

So how can we introduce semantic relation to few-shot detection? In natural language processing, semantic concepts are represented by word embeddings \cite{word2vec, glove} from language models, which have been used in zero-shot learning methods \cite{zsr-gnn, zsod}. And explicit relationships are represented by knowledge graphs \cite{wordnet, nell}, which are adopted by some zero-shot or few-shot recognition algorithms \cite{zsr-gnn, fsr-kt}. However, these techniques are rarely explored in the FSOD task. Also, directly applying them to few-shot detectors leads to non-trivial practical problems, i.e. the domain gap between vision and language, and the heuristic definition of knowledge graph for classes in FSOD datasets (see Section \ref{subsec:ssp} and \ref{subsec:rr} for details).

In this work, we explore the semantic relation for FSOD. We propose a Semantic Relation Reasoning Few-Shot Detector (SRR-FSD), which learns novel objects from both the visual information and the semantic relation in an end-to-end style. Specifically, we construct a semantic space using the word embeddings. Guided by the word embeddings of the classes, the detector is trained to project the objects from the visual space to the semantic space and to align their image representations with the corresponding class embeddings. To address the aforementioned problems, we propose to learn a dynamic relation graph driven by the image data instead of pre-defining one based on heuristics. Then the learned graph is used to perform relation reasoning and augment the raw embeddings for reduced domain gap.

With the help of the semantic relation reasoning, our SRR-FSD demonstrates the shot-stable property in two aspects, see the red solid and dashed lines in Figure \ref{fig:vocnov1}. In the common few-shot settings (solid lines), SRR-FSD achieves competitive performance at higher shots and significantly better performance at lower shots compared to state-of-the-art few-shot detectors. In a more realistic setting (dashed lines) where implicit shots of novel concepts are removed from the classification dataset for the pretrained model, SRR-FSD steadily maintains the performance while some previous methods have results degraded by a large margin due to the loss of implicit shots. We hope the suggested realistic setting can serve as a new benchmark protocol for future research. 

We summarize our contributions as follows:
\begin{itemize}
    \item To our knowledge, our work is the first to investigate semantic relation reasoning for the few-shot detection task and show its potential to improve a strong baseline.
    \item Our SRR-FSD achieves stable performance w.r.t the shot variation, outperforming state-of-the-art FSOD methods under several existing settings especially when the novel class data is extremely limited.
    \item We suggest a more realistic FSOD setting in which implicit shots of novel classes are removed from the classification dataset for the pretrained model, and show that our SRR-FSD can maintain a more steady performance compared to previous methods if using the new pretrained model. 
\end{itemize}

\section{Related Work}

\textbf{Object Detection}
Object detection is a fundamental computer vision task, serving as a necessary step for various down-streaming instance-based understanding. Modern CNN-based detectors can be roughly divided into two categories. One is single-stage detector such as YOLO \cite{yolov2}, SSD \cite{ssd}, RetinaNet \cite{retinanet}, and FreeAnchor \cite{freeanchor} which directly predict the class confidence scores and the bounding box coordinates over a dense grid. The other is multi-stage detector such as Faster R-CNN \cite{faster-rcnn}, R-FCN \cite{r-fcn}, FPN \cite{fpn}, Cascade R-CNN \cite{cascade-rcnn}, and Libra R-CNN \cite{libra-rcnn} which predict class-agnostic regions of interest and refine those region proposals for one or multiple times. All these methods rely on pre-defined anchor boxes to have an initial estimation of the size and aspect ratio of the objects. Recently, anchor-free detectors eliminate the performance-sensitive hyperparameters for the anchor design. Some of them detect the key points of bounding boxes \cite{cornernet, extremenet, centernet}. Some of them encode and decode the bounding boxes as anchor points and point-to-boundary distances \cite{guidedanchor, fsaf, fcos, rpdet, sapd}. DETR \cite{detr} reformulates object detection as a direct set prediction problem and solve it with transformers. However, these detectors are trained with full supervision where each class has abundant annotated object instances.

\textbf{Few-Shot Detection} 
Recently, there have been works focusing on solving the detection problem in the limited data scenario. LSTD \cite{lstd} proposes the transfer knowledge regularization and background depression regularization to promote the knowledge transfer from the source domain to the target domain. \cite{fsod-mc} proposes to iterate between model training and high-confidence sample selection. RepMet \cite{repmet} adopts a distance metric learning classifier into the RoI classification head. FSRW \cite{yolo-fewshot} and Meta R-CNN \cite{meta-rcnn} predict per-class attentive vectors to reweight the feature maps of the corresponding classes. MetaDet \cite{metadet} leverages meta-level knowledge about model parameter generation for category-specific components of novel classes. In \cite{attention-rpn}, the similarity between the few shot support set and query set is explored to detect novel objects. Context-Transformer \cite{context-transformer} relies on discriminative context clues to reduce object confusion. TFA \cite{tfa} only fine-tunes the last few layers of the detector. Two very recent papers are MPSR \cite{mpsr} and FSDetView \cite{fsdetview}. MPSR develops an auxiliary branch to generate multi-scale positive samples as object pyramids and to refine the prediction at various scales. FSDetView proposes a joint feature embedding module to share the feature from base classes. However, all these methods \textit{depend purely on visual information} and suffer from shot variation. 

\textbf{Semantic Reasoning in Vision Tasks}
Semantic word embeddings have been used in zero-shot learning tasks to learn a mapping from the visual feature space to the semantic space, such as zero-shot recognition \cite{zsr-gnn} and zero-shot object detection \cite{zsod, polarity}. In \cite{chen_etal}, semantic embeddings are used as the ground-truth of the encoder TriNet to guide the feature augmentation. In \cite{lu_etal}, semantic embeddings guide the feature synthesis for unseen classes by perturbing the seen feature with the projected difference between a seen class embedding and a unseen class embedding. In zero-shot or few-shot recognition \cite{zsr-gnn, fsr-kt}, word embeddings are often combined with knowledge graphs to perform relation reasoning via the graph convolution operation \cite{gcn}. Knowledge graphs are usually defined based on heuristics from databases of common sense knowledge rules \cite{wordnet, nell}. \cite{multi-label-gcn} proposed a knowledge graph based on object co-occurrence for the multi-label recognition task. To our knowledge, the use of word embeddings and knowledge graphs are rarely explored in the FSOD task. Any-Shot Detector (ASD) \cite{anyshot} is the only work that uses word embeddings for the FSOD task. But ASD focuses more on the zero-shot detection and it does not consider the explicit relation reasoning between classes because each word embedding is treated independently.

\section{Semantic Relation Reasoning Few-Shot Detector}
\label{sec:s2-fsd}
\begin{figure}
    \centering
    \includegraphics[width=\columnwidth]{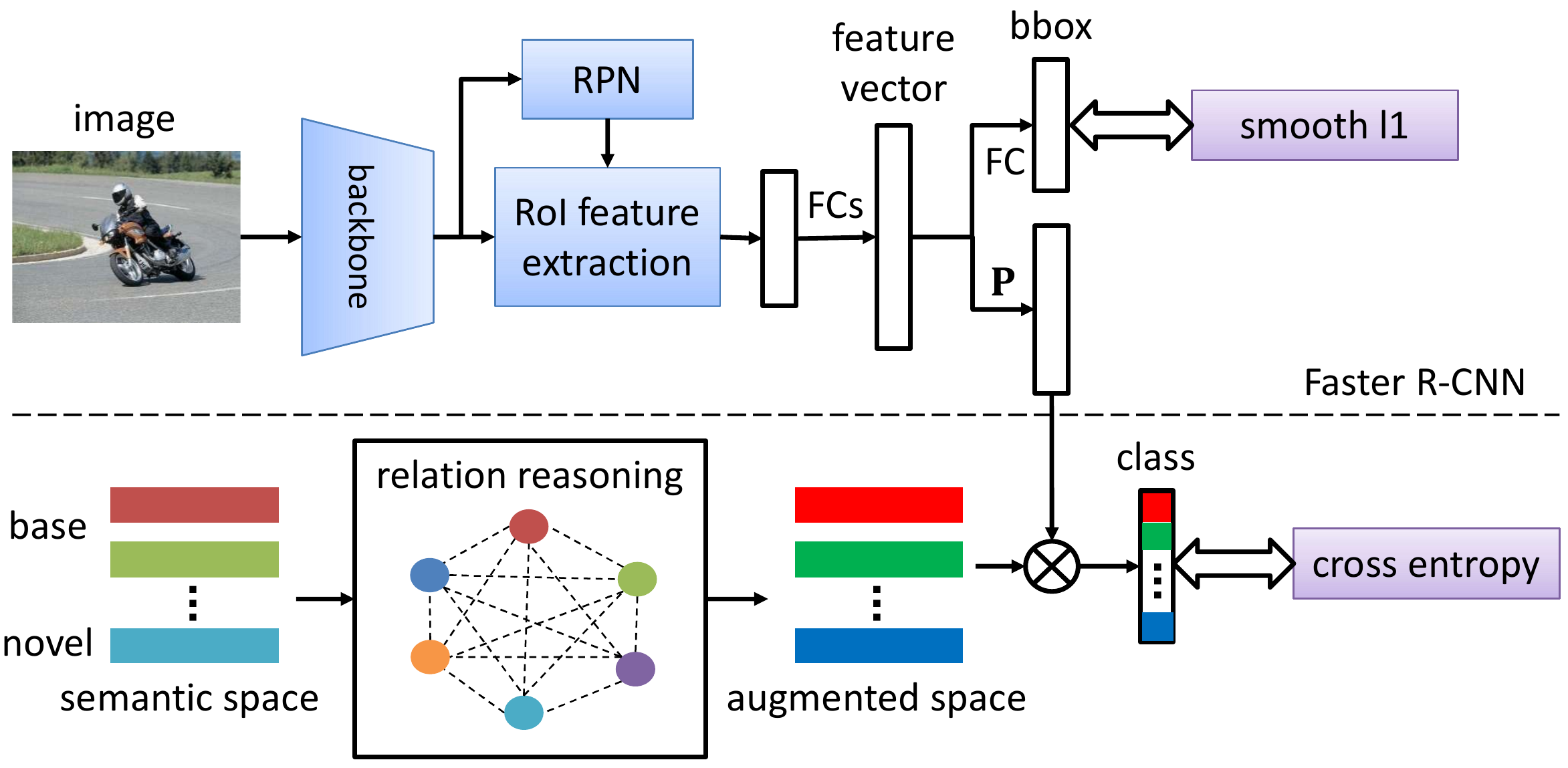}
    \caption{Overview of the SRR-FSD. A semantic space is built from the word embeddings of all corresponding classes in the dataset and is augmented through a relation reasoning module. Visual features are learned to be projected into the augmented space. ``$\bigotimes$'': dot product. ``FC'': fully-connected layer. ``$\mathbf{P}$'': lernable projection matrix.}
    \label{fig:s2-fsd}
\end{figure}

In this section, we first briefly introduce the preliminaries for few-shot object detection including the problem setup and the general training pipelines. Then based on Faster R-CNN \cite{faster-rcnn}, we build our SRR-FSD by integrating semantic relation with the visual information and allowing it to perform relation reasoning in the semantic space. We also discuss the problems of trivially using the raw word embeddings and the predefined knowledge graphs. Finally, we introduce the two-phase training processes. An overview of our SRR-FSD is illustrated in Figure \ref{fig:s2-fsd}.

\subsection{FSOD Preliminaries}
Conventional object detection problem has a base class set $\mathcal{C}_b$ in which there are many instances, and a base dataset $\mathcal{D}_b$ with abundant images. $\mathcal{D}_b$ consists of a set of annotated images $\{(x_i, y_i)\}$ where $x_i$ is the image and $y_i$ is the annotation of labels from $\mathcal{C}_b$ and bounding boxes for objects in $x_i$. For few-shot object detection (FSOD) problem, in addition to $\mathcal{C}_b$ and $\mathcal{D}_b$ it also has a novel class set $\mathcal{C}_n$ and a novel dataset $\mathcal{D}_n$, with $\mathcal{C}_b \cap \mathcal{C}_n = \emptyset$. In $\mathcal{D}_n$, objects have labels belong to $\mathcal{C}_n$ and the number of objects for each class is $k$ for $k$-shot detection. A few-shot detector is expected to learn from $\mathcal{D}_b$ and to quickly generalize to $\mathcal{D}_n$ with a small $k$ such that it can detect all objects in a held-out testing set with object classes in $\mathcal{C}_b \cup \mathcal{C}_n$. We assume all classes in $\mathcal{C}_b \cup \mathcal{C}_n$ have semantically meaningful names so the corresponding semantic embeddings can be retrieved.

A typical few-shot detector has two training phases. The first one is the base training phase where the detector is trained on $\mathcal{D}_b$ similarly to conventional object detectors. Then in the second phase, it is further fine-tuned on the union of $\mathcal{D}_b$ and $\mathcal{D}_n$. To avoid the dominance of objects from $\mathcal{D}_b$, a small subset is sampled from $\mathcal{D}_b$ such that the training set is balanced concerning the number of objects per class. As the total number of classes is increased by the size of $\mathcal{C}_n$ in the second phase, more class-specific parameters are inserted in the detector and trained to be responsible for the detection of novel objects. The class-specific parameters are usually in the box classification and localization layers at the very end of the network.

\subsection{Semantic Space Projection}
\label{subsec:ssp}
Our few-shot detector is built on top of Faster R-CNN \cite{faster-rcnn}, a popular two-stage general object detector. In the second-stage of Faster R-CNN, a feature vector is extracted for each region proposal and forwarded to a classification subnet and a regression subnet. In the classification subnet, the feature vector is transformed into a $d$-dimentional vector $\mathbf{v} \in \mathcal{R}^d$ through fully-connected layers. Then $\mathbf{v}$ is multiplied by a learnable weight matrix $\mathbf{W} \in \mathcal{R}^{N \times d}$ to output a probability distribution as in Eq. \eqref{eq:softmax}.
\begin{equation}
\label{eq:softmax}
    \mathbf{p} = softmax(\mathbf{W}\mathbf{v} + \mathbf{b})
\end{equation}
where $N$ is the number of classes and $\mathbf{b} \in \mathcal{R}^N$ is a learnable bias vector. Cross-entropy loss is used during training.

To learn objects from both the visual information and the semantic relation, we first construct a semantic space and project the visual feature $\mathbf{v}$ into this semantic space. Specifically, we represent the semantic space using a set of $d_e$-dimensional word embeddings $\mathbf{W}_e \in \mathcal{R}^{N \times d_e}$ \cite{word2vec} corresponding to the $N$ object classes (including the background class). And the detector is trained to learn a linear projection $\mathbf{P} \in \mathcal{R}^{d_e \times d}$ in the classification subnet (see Figure \ref{fig:s2-fsd}) such that $\mathbf{v}$ is expected to align with its class's word embedding after projection. Mathematically, the prediction of the probability distribution turns into Eq. \eqref{eq:projection} from Eq. \eqref{eq:softmax}.
\begin{equation}
\label{eq:projection}
\mathbf{p} = softmax(\mathbf{W}_e \mathbf{P} \mathbf{v} + \mathbf{b})
\end{equation}
During training, $\mathbf{W}_e$ is fixed and the learnable variable is $\mathbf{P}$. A benefit is that generalization to novel objects involves no new parameters in $\mathbf{P}$. We can simply expand $\mathbf{W}_e$ with embeddings of novel classes. We still keep the $\mathbf{b}$ to model the category imbalance in the detection dataset.

\textbf{Domain gap between vision and language.} $\mathbf{W}_e$ encodes the knowledge of semantic concepts from natural language. While it is applicable in zero-shot learning, it will introduce the bias of the domain gap between vision and language to the FSOD task. Because unlike zero-shot learning where unseen classes have no support from images, the few-shot detector can rely on both the images and the embeddings to learn the concept of novel objects. When there are very few images to rely on, the knowledge from embeddings can guide the detector towards a decent solution. But when more images are available, the knowledge from embeddings may be misleading due to the domain gap, resulting in a suboptimal solution. Therefore, we need to augment the semantic embeddings to reduce the domain gap. Some previous works like ASD \cite{anyshot} apply a trainable transformation to each word embedding \textit{independently}. But we leveraging the explicit relationship between classes is more effective for embedding augmentation, leading to the proposal of the dynamic relation graph in Section \ref{subsec:rr}.

\subsection{Relation Reasoning}
\label{subsec:rr}

The semantic space projection learns to align the concepts from the visual space with the semantic space. But it still treats each class independently and there is no knowledge propagation among classes. Therefore, we further introduce a knowledge graph to model their relationships. The knowledge graph $\mathbf{G}$ is a $N \times N$ adjacency matrix representing the connection strength for every neighboring class pairs. $\mathbf{G}$ is involved in classification via the graph convolution operation \cite{gcn}. Mathematically, the updated probability prediction is shown in Eq. \eqref{eq:graph}.
\begin{equation}
\label{eq:graph}
\mathbf{p} = softmax(\mathbf{G} \mathbf{W}_e \mathbf{P} \mathbf{v} + \mathbf{b})
\end{equation}

\begin{figure}
    \centering
    \includegraphics[width=\columnwidth]{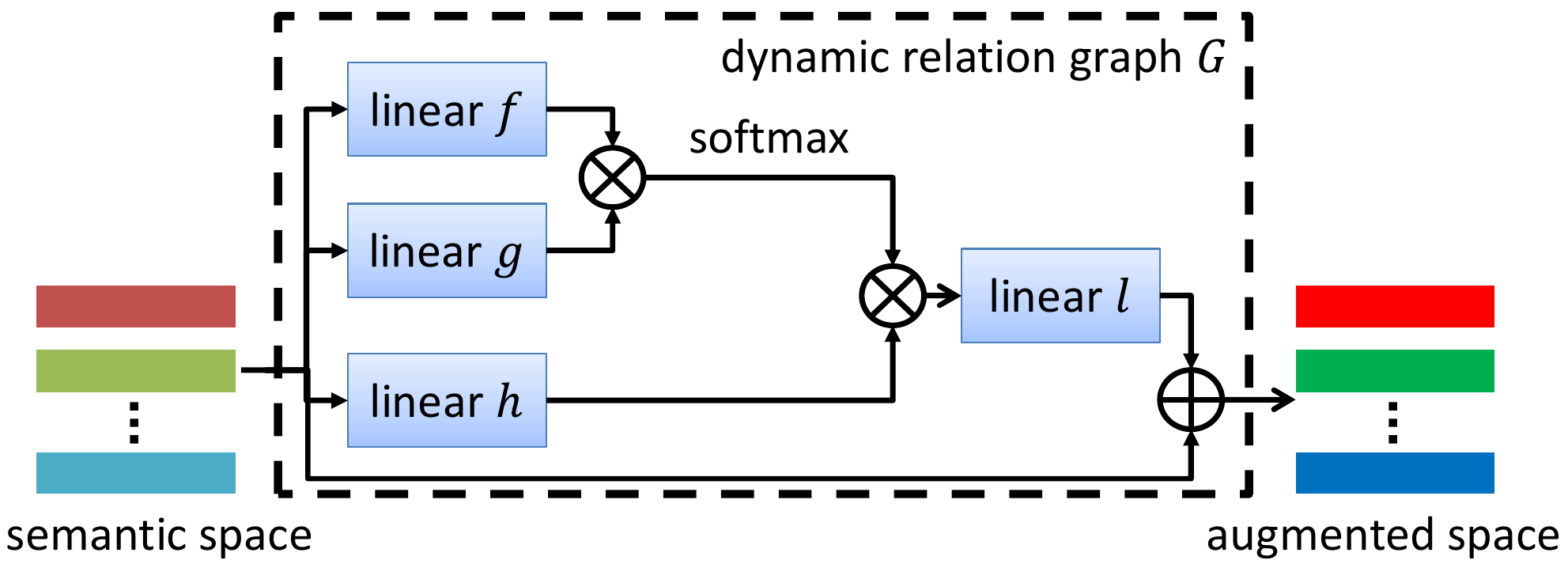}
    \caption{Network architecture of the relation reasoning module for learning the relation graph. ``$\bigotimes$'': dot product. ``$\bigoplus$'': element-wise plus.}
    \label{fig:relation_reasoning}
\end{figure}

\textbf{The heuristic definition of the knowledge graph.} In zero-shot or few-shot recognition algorithms, the knowledge graph $\mathbf{G}$ is predefined base on heuristics. It is usually constructed from a database of common sense knowledge rules by sampling a sub-graph through the rule paths such that semantically related classes have strong connections. For example, classes from the ImageNet dataset \cite{imagenet} have a knowledge graph sampled from the WordNet \cite{wordnet}. However, classes in FSOD datasets are not highly semantically related, nor do they form a hierarchical structure like the ImageNet classes. The only applicable heuristics we found are based on object co-occurrence from \cite{multi-label-gcn}. Although the statistics of the co-occurrence are straightforward to compute, the co-occurrence is not necessarily equivalent to semantic relation.

Instead of predefining a knowledge graph based on heuristics, we propose to learn a \textit{dynamic} relation graph driven by the data to model the relation reasoning between classes. The data-driven graph is also responsible for reducing the domain gap between vision and language because it is trained with image inputs. Inspired by the concept of the transformer, we implement the dynamic graph with the self-attention architecture \cite{attention} as shown in Figure \ref{fig:relation_reasoning}. The original word embeddings $\mathbf{W}_e$ are transformed by three linear layers $f, g, h$, and a self-attention matrix is computed from the outputs of $f, g$. The self-attention matrix is multiplied with the output of $h$ followed by another linear layer $l$. A residual connection \cite{resnet} adds the output of $l$ with the original $\mathbf{W}_e$. Another advantage of learning a dynamic graph is that it can easily adapt to new coming classes. Because the graph is not fixed and is generated on the fly from the word embeddings. We do not need to redefine a new graph and retrain the detector from the beginning. We can simply insert corresponding embeddings of new classes and fine-tune the detector.


\subsection{Decoupled Fine-tuning}
\label{subsec:df}
In the second fine-tuning phase, we only unfreeze the last few layers of our SRR-FSD similar to TFA \cite{tfa}. For the classification subnet, we fine-tune the parameters in the relation reasoning module and the projection matrix $\mathbf{P}$. For the localization subnet, it is not dependent on the word embeddings but it shares features with the classification subnet. We find that the learning of localization on novel objects can interfere with the classification subnet via the shared features, leading to many false positives. Decoupling the shared fully-connected layers between the two subnets can effectively make each subnet learn better features for its task. In other words, the classification subnet and the localization subnet have individual fully-connected layers and they are fine-tuned independently.

\section{Experiments}

\subsection{Implementation Details}
Our SRR-FSD is implemented based on Faster R-CNN \cite{faster-rcnn} with ResNet-101 \cite{resnet} and Feature Pyramid Network \cite{fpn} as the backbone using the MMDetection \cite{mmdet} framework. All models are trained with Stochastic Gradient Descent (SGD) and a batch size of 16. For the word embeddings, we use the L2-normalized 300-dimensional Word2Vec \cite{word2vec} vectors from the language model trained on large unannotated texts like Wikipedia. In the relation reasoning module, we reduce the dimension of word embeddings to 32 which is empirically selected. In the first base training phase, we set the learning rate, the momentum, and the weight decay to 0.02, 0.9, and 0.0001, respectively. In the second fine-tuning phase, we reduce the learning rate to 0.001 unless otherwise mentioned. The input image is sampled by first randomly choosing between the base set and the novel set with a 50\% probability and then randomly selecting an image from the chosen set.

\subsection{Existing Settings}
We follow the existing settings in previous FSOD methods \cite{yolo-fewshot,metadet,meta-rcnn,tfa} to evaluate our SRR-FSD on the VOC \cite{voc} and COCO \cite{coco} datasets. For fair comparison and reduced randomness, we use the same data splits and a fixed list of novel samples provided by \cite{yolo-fewshot}.

\begin{table*}
\centering
\setlength\tabcolsep{5pt}
\begin{tabular}{c|ccccc|ccccc|ccccc}
\hline \hline
                     & \multicolumn{5}{c|}{Novel Set 1} & \multicolumn{5}{c|}{Novel Set 2} & \multicolumn{5}{c}{Novel Set 3} \\
Method / shot        & 1     & 2    & 3    & 5    & 10    & 1     & 2    & 3    & 5    & 10    & 1     & 2    & 3    & 5    & 10    \\ \hline
FSRW \cite{yolo-fewshot} & 14.8 & 15.5 & 26.7 & 33.9 & 47.2 & 15.7 & 15.3 & 22.7 & 30.1 & 40.5 & 21.3 & 25.6 & 28.4 & 42.8 & 45.9 \\
MetaDet \cite{metadet} & 18.9 & 20.6 & 30.2 & 36.8 & 49.6 & 21.8 & 23.1 & 27.8 & 31.7 & 43.0 & 20.6 & 23.9 & 29.4 & 43.9 & 44.1 \\
Meta R-CNN \cite{meta-rcnn}     & 19.9 & 25.5 & 35.0 & 45.7 & 51.5 & 10.4 & 19.4 & 29.6 & 34.8 & \textbf{45.4} & 14.3 & 18.2 & 27.5 & 41.2 & 48.1 \\ 
TFA \cite{tfa} & 39.8 & 36.1 & 44.7 & \textbf{55.7} & 56.0 & 23.5 & 26.9 & 34.1 & 35.1 & 39.1 & 30.8 & 34.8 & 42.8 & \textbf{49.5} & \textbf{49.8} \\
\hline
SRR-FSD (Ours) & \textbf{47.8} & \textbf{50.5} & \textbf{51.3} & 55.2 & \textbf{56.8} & \textbf{32.5} & \textbf{35.3} & \textbf{39.1} & \textbf{40.8} & 43.8 & \textbf{40.1} & \textbf{41.5} & \textbf{44.3} & 46.9 & 46.4 \\ \hline
\end{tabular}
\caption{FSOD evaluation on VOC. We report the mAP with IoU threshold 0.5 (mAP50) under 3 different sets of 5 novel classes with
a small number of shots.}
\label{tab:voc}
\end{table*}

\textbf{VOC} The 07 and 12 train/val sets are used for training and the 07 test set is for testing. Out of its 20 object classes, 5 classes are selected as novel and the remaining 15 are base classes, with 3 different base/novel splits. The novel classes each have $k$ annotated objects, where $k$ equals 1, 2, 3, 5, 10. In the first base training phase, our SRR-FSD is trained for 18 epochs with the learning rate multiplied by 0.1 at the 12th and 15th epoch. In the second fine-tuning phase, we train for $500 \times |\mathcal{D}_n|$ steps where $|\mathcal{D}_n|$ is the number of images in the $k$-shot novel dataset.

We report the mAP50 of the novel classes on VOC with 3 splits in Table \ref{tab:voc}. In all different base/novel splits, our SRR-FSD achieves a more shot-stable performance. At higher shots like 5-shot and 10-shot, our performance is competitive compared to previous state-of-the-art methods. At more challenging conditions with shots less than 5, our approach can outperform the second-best by a large margin (up to 10+ mAP). Compared to ASD \cite{anyshot} which only reports results of 3-shot and 5-shot in the Novel Set 1, ours is 24.2 and 6.0 better respectively in mAP. We do not include ASD in Table \ref{tab:voc} because its paper does not provide the complete results on VOC.

\textit{Learning without forgetting} is another merit of our SRR-FSD. After generalization to novel objects, the performance on the base objects does not drop at all as shown in Table \ref{tab:vocbase+nov}.  Both base AP and novel AP of our SRR-FSD compare favorably to previous methods based on the same Faster R-CNN with ResNet-101. The base AP even increases a bit probably due to the semantic relation reasoning from limited novel objects to base objects.

\begin{table}
\centering
\begin{tabular}{c|c|cc}
\hline \hline
Shot               & Method        & Base AP50 & Novel AP50 \\ \hline
\multirow{4}{*}{3}  & Meta R-CNN \cite{meta-rcnn}    & 64.8      & 35.0       \\
                    & TFA \cite{tfa}          & 79.1      & 44.7       \\
                    & Ours base only & 77.7 & n/a \\
                    & SRR-FSD (Ours) & 78.2      & 51.3       \\ \hline
\multirow{4}{*}{10} & Meta R-CNN \cite{meta-rcnn}    & 67.9      & 51.5       \\
                    & TFA \cite{tfa}          & 78.4      & 56.0       \\
                    & Ours base only & 77.7 & n/a \\
                    & SRR-FSD (Ours) & 78.2      & 56.8       \\ \hline
\end{tabular}
\caption{FSOD performance for the base and novel
classes on Novel Set 1 of VOC. Our SRR-FSD has the merit of learning without forgetting.}
\label{tab:vocbase+nov}
\end{table}

\textbf{COCO} The \texttt{minival} set with 5000 images is used for testing and the rest images in train/val sets are for training. Out of the 80 classes, 20 of them overlapped with VOC are the novel classes with $k=10, 30$ shots per class and the remaining 60 classes are base. We train the SRR-FSD on the base dataset for 12 epochs using the same setting as MMDetection \cite{mmdet} and fine-tune it for a fixed number of $10 \times |\mathcal{D}_b|$ steps where $|\mathcal{D}_b|$ is the number of images in the base dataset. Unlike VOC, the base dataset in COCO contains unlabeled novel objects, so the region proposal network (RPN) treats them as the background. To avoid omitting novel objects in the fine-tuning phase, we unfreeze the RPN and the following layers.
Table \ref{tab:coco} presents the COCO-style averaged AP. Again we consistently outperform previous methods including FSRW \cite{yolo-fewshot}, MetaDet \cite{metadet}, Meta R-CNN \cite{meta-rcnn}, TFA \cite{tfa}, and MPSR \cite{mpsr}.

\begin{table}
\centering
\begin{tabular}{c|c|ccc}
\hline \hline
Shot                & Method              & AP & AP50 & AP75 \\ \hline
\multirow{6}{*}{10} & FSRW \cite{yolo-fewshot}      & 5.6 & 12.3 & 4.6 \\
& MetaDet \cite{metadet} & 7.1 & 14.6 & 6.1 \\
& Meta R-CNN \cite{meta-rcnn}    & 8.7 & 19.1 & 6.6 \\
& TFA \cite{tfa} & 10.0 & - & 9.3 \\
& MPSR \cite{mpsr} & 9.8 & 17.9 & 9.7 \\
& SRR-FSD (Ours) & \textbf{11.3} & \textbf{23.0} & \textbf{9.8} \\
\hline
\multirow{6}{*}{30} & FSRW \cite{yolo-fewshot} & 9.1 & 19.0 & 7.6 \\
& MetaDet \cite{metadet} & 11.3 & 21.7 & 8.1 \\
& Meta R-CNN \cite{meta-rcnn} & 12.4 & 25.3 & 10.8 \\
& TFA \cite{tfa} & 13.7 & - & 13.4 \\
& MPSR \cite{mpsr} & 14.1 & 25.4 & \textbf{14.2} \\
& SRR-FSD (Ours) & \textbf{14.7} & \textbf{29.2} & 13.5 \\
\hline
\end{tabular}
\caption{FSOD performance of the novel classes on COCO.}
\label{tab:coco}
\end{table}

\begin{figure}
    \centering
    \includegraphics[width=0.9\columnwidth]{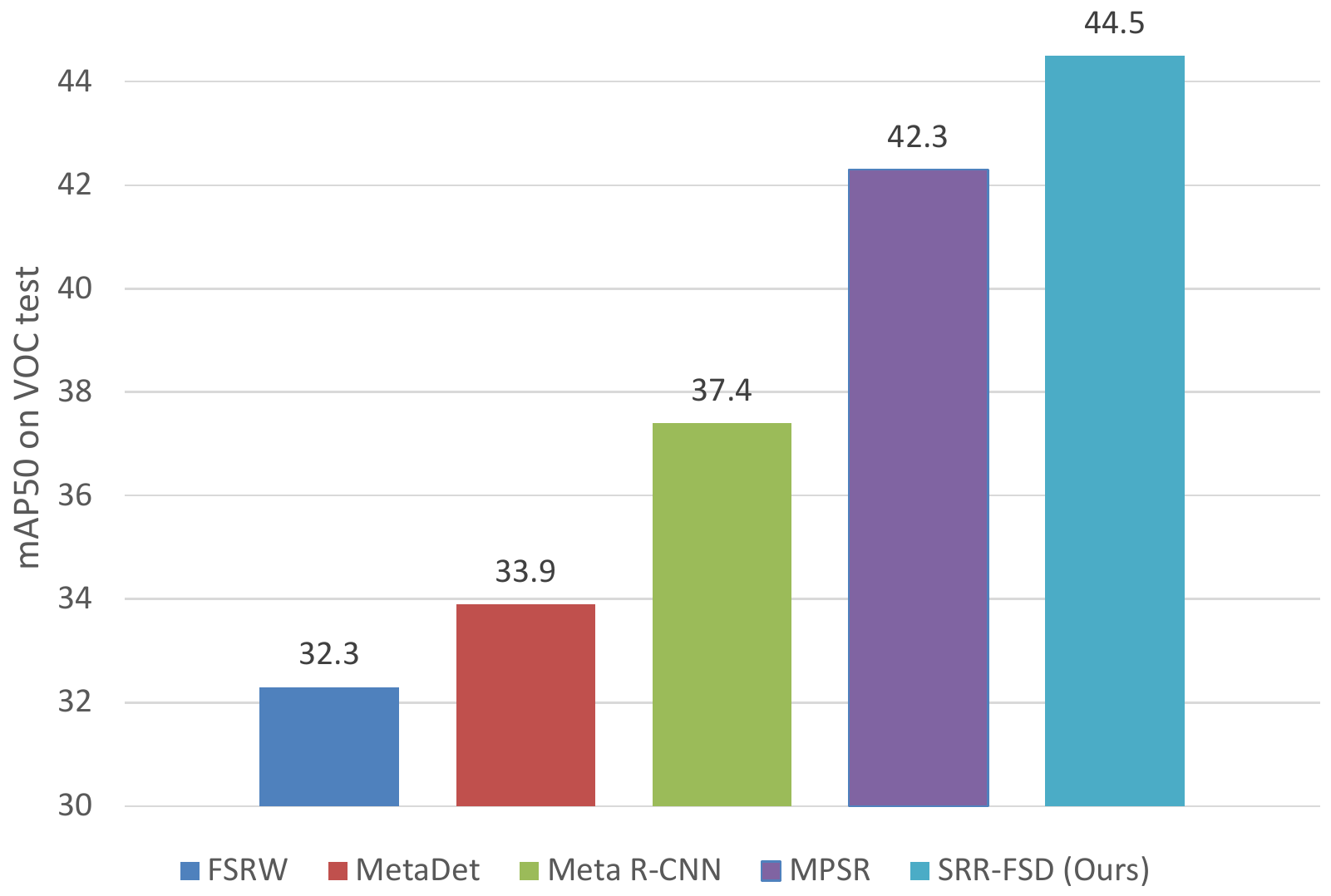}
    \caption{10-shot cross domain performance on the 20 novel classes under COCO to VOC.}
    \label{fig:coco2voc}
\end{figure}

\textbf{COCO to VOC} For the cross-domain FSOD setting, we follow \cite{yolo-fewshot, metadet} to use the same base dataset with 60 classes as in the previous COCO within-domain setting. The novel dataset consists of 10 samples for each of the 20 classes from the VOC dataset. The learning schedule is the same as the previous COCO within-domain setting except the learning rate is 0.005. Figure \ref{fig:coco2voc} shows that our SRR-FSD achieves the best performance with a healthy 44.5 mAP, indicating better generalization ability in cross-domain situations.

\subsection{A More Realistic Setting}

\begin{table*}
\centering
\setlength\tabcolsep{5pt}
\begin{tabular}{c|ccccc|ccccc|ccccc}
\hline \hline
                     & \multicolumn{5}{c|}{Novel Set 1} & \multicolumn{5}{c|}{Novel Set 2} & \multicolumn{5}{c}{Novel Set 3} \\
Method / shot        & 1     & 2    & 3    & 5    & 10    & 1     & 2    & 3    & 5    & 10    & 1     & 2    & 3    & 5    & 10    \\ \hline
FSRW \cite{yolo-fewshot} & 13.9 & 21.1 & 20.0 & 29.9 & 40.8 & 13.5 & 14.2 & 20.6 & 20.7 & 36.8 & 16.2 & 22.2 & 26.8 & 37.0 & 41.5 \\ 
Meta R-CNN \cite{meta-rcnn} & 11.5 & 22.2 & 24.7 & 36.4 & 45.2 & 10.1 & 16.9 & 22.7 & 29.6 & 40.1 & 10.0 & 21.7 & 27.1 & 32.8 & 41.6 \\ 
TFA \cite{tfa} & 35.8 & 39.5 & 44.2 & 50.8 & 55.3 & 18.8 & 26.0 & 33.2 & 31.3 & 39.2 & 25.6 & 32.6 & 36.4 & \textbf{43.7} & \textbf{48.5} \\ \hline
SRR-FSD (Ours) & \textbf{46.3} & \textbf{51.1} & \textbf{52.6} & \textbf{56.2} & \textbf{57.3} & \textbf{31.0} & \textbf{29.9} & \textbf{34.7} & \textbf{37.3} & \textbf{41.7} & \textbf{39.2} & \textbf{40.5} & \textbf{39.7} & 42.2 & 45.2 \\
\hline
\end{tabular}
\caption{FSOD performance (mAP50) on VOC under a more realistic setting where novel classes are removed from the pretrained classification dataset to guarantee $\mathcal{C}_0 \cap \mathcal{C}_n = \emptyset$. Our SRR-FSD is more robust to the loss of implicit shots comparing with Table \ref{tab:voc}.}
\label{tab:vocdenov}
\end{table*}

The training of the few-shot detector usually involves initializing the backbone network with a model pretrained on large-scale object classification datasets such as ImageNet \cite{imagenet}. The set of object classes in ImageNet, i.e. $\mathcal{C}_0$, is highly overlapped with the novel class set $\mathcal{C}_n$ in the existing settings. This means that the pretrained model can get early access to large amounts of object samples, i.e. \textit{implicit shots}, from novel classes and encode their knowledge in the parameters before it is further trained for the detection task. Even the pretrained model is optimized for the recognition task, the extracted features still have a big impact on the detection of novel objects (see Figure \ref{fig:vocnov1}). However, some rare classes may have highly limited or valuable data in the real world that pretraining a classification network on it is not realistic.

Therefore, we suggest a more realistic setting for FSOD, which extends the existing settings. In addition to $\mathcal{C}_b \cap \mathcal{C}_n = \emptyset$, we also require that $\mathcal{C}_0 \cap \mathcal{C}_n = \emptyset$. To achieve this, we systematically and hierarchically remove novel classes from $\mathcal{C}_0$. For each class in $\mathcal{C}_n$, we find its corresponding synset in ImageNet and obtain its full hyponym (the synset of the whole subtree starting from that synset) using the ImageNet API \footnote{http://image-net.org/download-API}. The images of this synset and its full hyponym are removed from the pretrained dataset. And the classification model is trained on a dataset with no novel objects. We provide the list of WordNet IDs for each novel class to be removed in Appendix \ref{ap:real_setting}. 

We notice that CoAE \cite{coae} also proposed to remove all COCO-related ImageNet classes to ensure the model does not ``foresee'' the unseen classes. As a result, a total of 275 classes are removed from ImageNet including both the base and novel classes in VOC \cite{voc}, which correspond to more than 300k images. We think the loss of this much data may lead to a worse pretrained model in general. So the pretrained model may not be able to extract features strong enough for down-streaming vision tasks compared with the model trained on full ImageNet. Our setting, on the other hand, tries to alleviate this effect as much as possible by only removing the novel classes in VOC Novel Set 1, 2, and 3 respectively, which correspond to an average of 50 classes from ImageNet.

Under the new realistic setting, we re-evaluate previous methods using their official source code and report the performance on the VOC dataset in Table \ref{tab:vocdenov}. Our SRR-FSD demonstrates superior performance to other methods under most conditions, especially at challenging lower shot scenarios. More importantly, our SRR-FSD is less affected by the loss of implicit shots. Compared with results in Table \ref{tab:voc}, our performance is more stably maintained when novel objects are only available in the novel dataset.

\subsection{Ablation Study}
In this section, we study the contribution of each component. Experiments are conducted on the VOC dataset. Our baseline is the Faster R-CNN \cite{faster-rcnn} with ResNet-101 \cite{resnet} and FPN \cite{fpn}. We gradually apply the Semantic Space Projection (SSP \ref{subsec:ssp}), Relation Reasoning (RR \ref{subsec:rr}) and Decoupled Fine-tuning (DF \ref{subsec:df}) to the baseline and report the performance in Table \ref{tab:ablation}. We also compare three different ways of augmenting the raw word embeddings in Table \ref{tab:relation_graph}, including the trainable transformation from ASD \cite{anyshot}, the heuristic knowledge graph from \cite{multi-label-gcn}, and the dynamic graph from our proposed relation reasoning module.

\begin{table*}
\setlength\tabcolsep{10pt}
\centering
\begin{tabular}{c|ccc|ccccc}
\hline \hline
\multirow{2}{*}{} & \multicolumn{3}{c|}{Components} & \multicolumn{5}{c}{Shots in Novel Set 1} \\
                  & SSP         & RR         & DF         & 1     & 2     & 3     & 5     & 10    \\ \hline
Faster R-CNN \cite{faster-rcnn}     &             &            &            & 32.6  & 44.4  & 46.3  & 49.6  & 55.6  \\
                  & \checkmark  &            &            & 40.5  & 46.8  & 46.5  & 47.1  & 52.2  \\
                  & \checkmark  & \checkmark &            & 44.1  & 46.0  & 47.8  & 51.7  & 54.7  \\
SRR-FSD            & \checkmark  & \checkmark & \checkmark & 47.8  & 50.5  & 51.3  & 55.2  & 56.8  \\ \hline
\end{tabular}
\caption{Ablative performance (mAP50) on the VOC Novel Set 1 by gradually applying the proposed components to the baseline Faster R-CNN. \textbf{SSP}: semantic space projection. \textbf{RR}: relation reasoning. \textbf{DF}: decoupled fine-tuning.}
\label{tab:ablation}
\end{table*}

\textbf{Semantic space projection guides shot-stable learning.} The baseline Faster R-CNN can already achieve satisfying results at 5-shot and 10-shot. But at 1-shot and 2-shot, performance starts to fall apart due to exclusive dependence on images. The semantic space projection, on the other hand, makes the learning more stable to the variation of shot numbers (see 1st and 2nd entries in Table \ref{tab:ablation}). The space projection guided by the semantic embeddings is learned well enough in the base training phase so it can be quickly adapted to novel classes with a few instances. We can observe a major boost at lower shot conditions compared to baseline, i.e. 7.9 mAP and 2.4 mAP gain at 1-shot and 2-shot respectively. However, the raw semantic embeddings limit the performance at higher shot conditions. The performance at 5-shot and 10-shot drops below the baseline. This verifies our argument about the domain gap between vision and language. At lower shots, there is not much visual information to rely on so the language information can guide the detector to a decent solution. But when more images are available, the visual information becomes more precise then the language information starts to be misleading. Therefore, we propose to refine the word embeddings for a reduced domain gap.

\begin{table}
\centering
\begin{tabular}{c|ccccc}
\hline \hline
\multirow{2}{*}{}             & \multicolumn{5}{c}{Shots in Novel Set 1}                                                                                             \\
                              & 1                        & 2                        & 3                        & 5                        & 10                       \\ \hline
+SSP                          & 40.5                     & 46.8                     & 46.5                     & 47.1                     & 52.2                     \\
\multicolumn{1}{l|}{+SSP +TT \cite{anyshot}} & \multicolumn{1}{l}{39.3} & \multicolumn{1}{l}{45.7} & \multicolumn{1}{l}{43.9} & \multicolumn{1}{l}{49.4} & \multicolumn{1}{l}{52.4} \\
+SSP +HKG \cite{multi-label-gcn}                     & 41.6                     & 45.5                     & 47.8                     & 49.7                     & 52.5                     \\
+SSP +RR                      & 44.1                     & 46.0                     & 47.8                     & 51.7                     & 54.7                     \\ \hline
\end{tabular}
\caption{Comparison of three ways of refining the word embeddings, including the trainable transformation from ASD \cite{anyshot}, the heuristic knowledge graph from \cite{multi-label-gcn}, and the dynamic relation graph from our relation reasoning module. \textbf{SSP}: semantic space projection. \textbf{RR}: relation reasoning. \textbf{TT}: trainable transformation. \textbf{HKG}: heuristic knowledge graph.}
\label{tab:relation_graph}
\end{table}


\textbf{Relation reasoning promotes adaptive knowledge propagation.} The relation reasoning module explicitly learns a relation graph that builds direct connections between base classes and novel classes. So the detector can learn the novel objects using the knowledge of base objects besides the visual information. Additionally, the relation reasoning module also functions as a refinement to the raw word embeddings with a data-driven relation graph. Since the relation graph is updated with image inputs, the refinement tends to adapt the word embeddings for the vision domain.
Results in Table \ref{tab:ablation} (2nd and 3rd entries) confirm that applying relation reasoning improves the detection accuracy of novel objects under different shot conditions. We also compare it with two other ways of refining the raw word embeddings in Table \ref{tab:relation_graph}. One is the trainable transformation (TT) from ASD \cite{anyshot} where word embeddings are updated with a trainable metric and a word vocabulary. Note that this transformation is applied to each embedding independently which does not consider the explicit relationships between them. The other one is the heuristic knowledge graph (HKG) defined based on the co-occurrence of objects from \cite{multi-label-gcn}. It turns out both the trainable transformation and the predefined heuristic knowledge graph are not as effective as the dynamic relation graph in the relation reasoning module. The effect of the trainable transformation is similar to unfreezing more parameters of the last few layers during fine-tuning as shown in Appendix \ref{ap:finetune}, which leads to overfitting when the shot is low. And the predefined knowledge graph is fixed during training thus cannot be adaptive to the inputs. In other words, the dynamic relation graph is better because it can not only perform explicit relation reasoning but also augment the raw embeddings for reduced domain gap between vision and language.

\begin{figure}
    \centering
    \begin{subfigure}
        \centering
        \rotatebox[origin=l]{90}{w/o DF}
        \includegraphics[width=0.17\columnwidth]{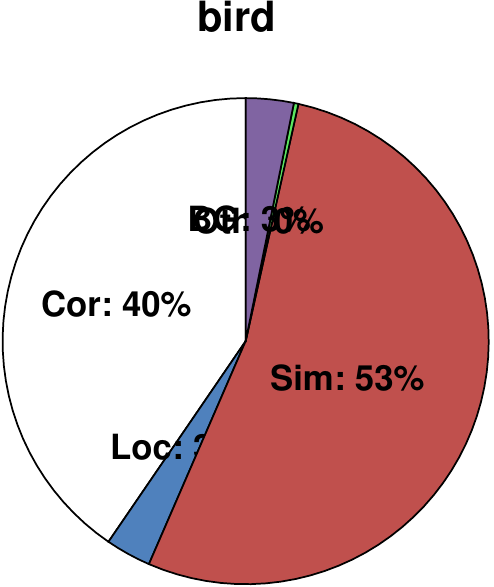}
        \includegraphics[width=0.17\columnwidth]{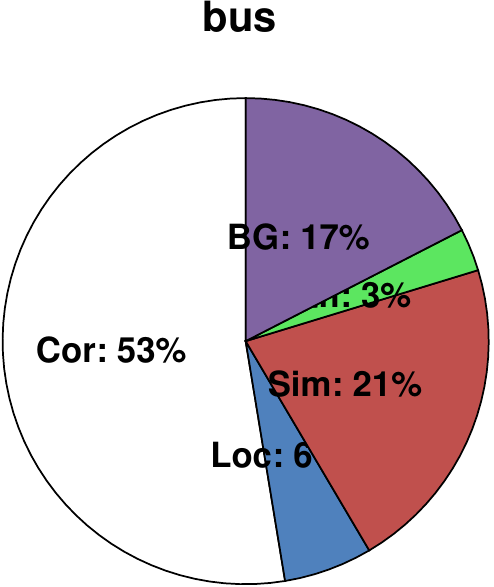}
        \includegraphics[width=0.17\columnwidth]{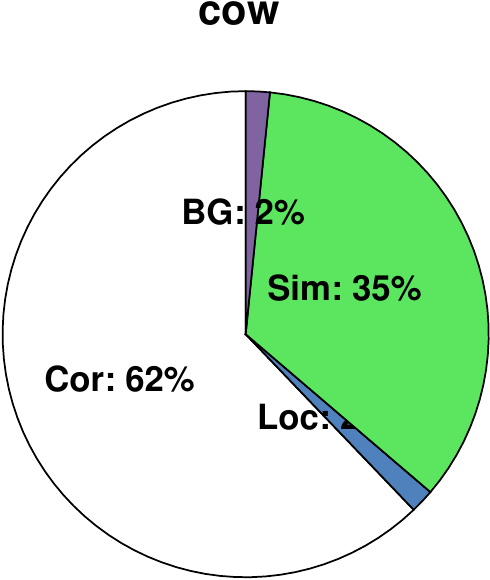}
        \includegraphics[width=0.17\columnwidth]{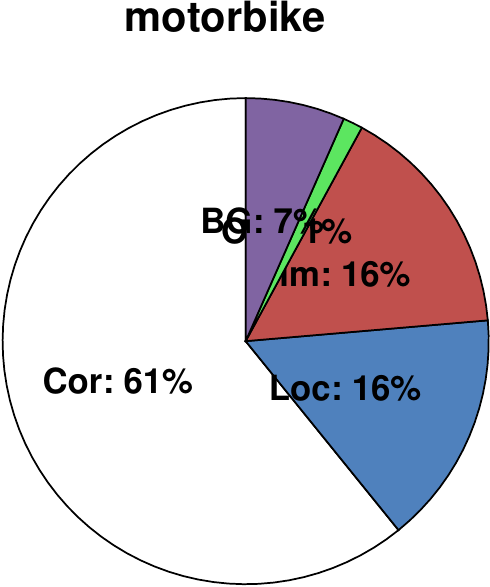}
        \includegraphics[width=0.17\columnwidth]{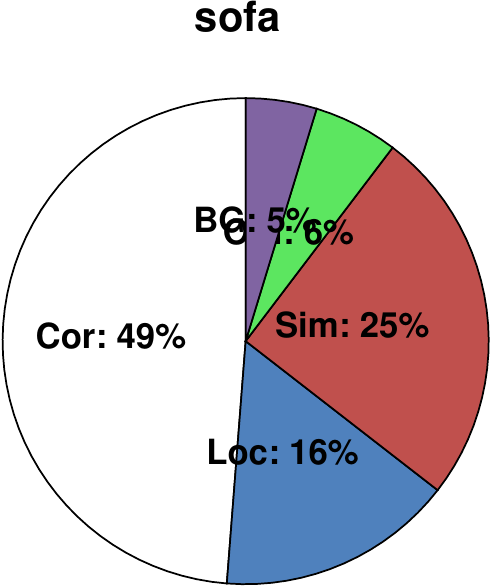}
    \end{subfigure}
    \hspace{4cm}
    \begin{subfigure}
        \centering
        \rotatebox[origin=l]{90}{w/ DF}
        \includegraphics[width=0.17\columnwidth]{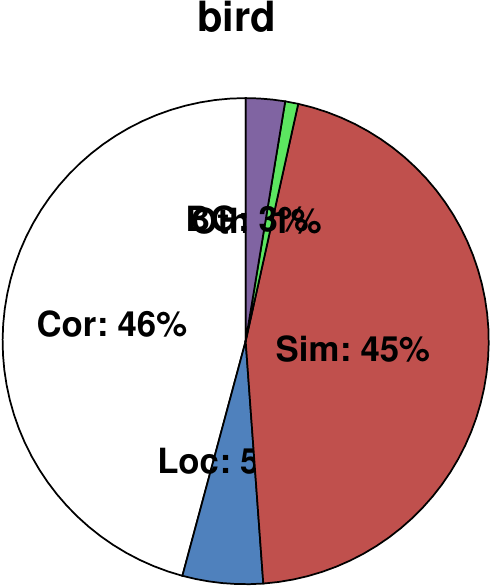}
        \includegraphics[width=0.17\columnwidth]{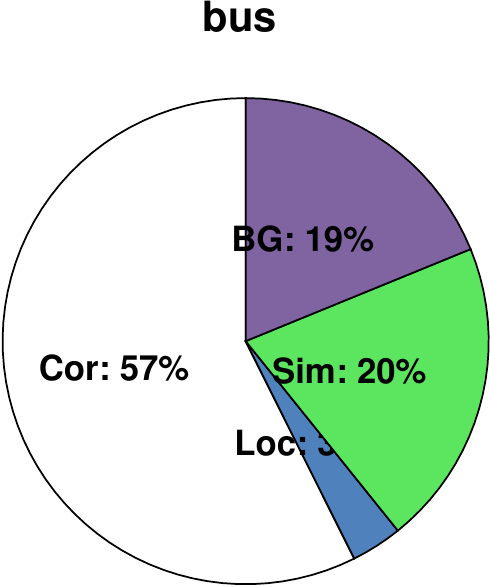}
        \includegraphics[width=0.17\columnwidth]{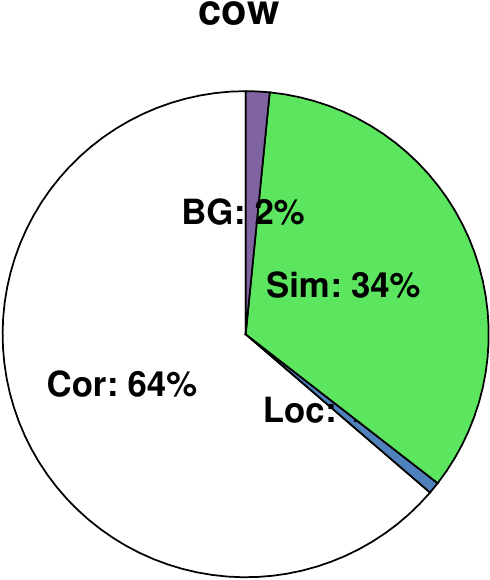}
        \includegraphics[width=0.17\columnwidth]{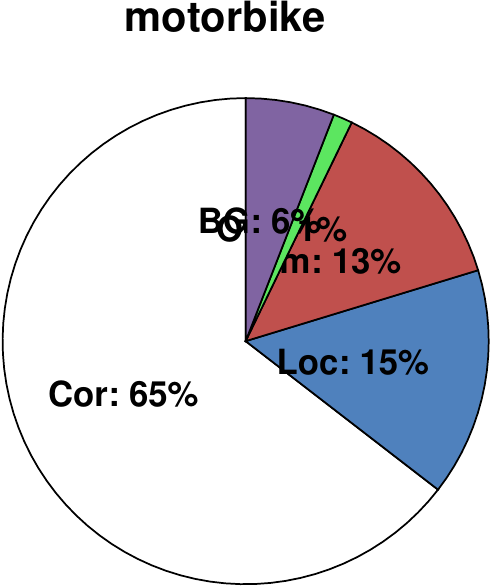}
        \includegraphics[width=0.17\columnwidth]{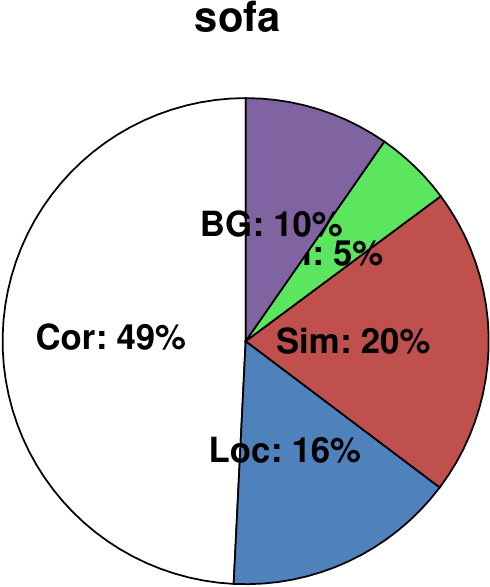}
    \end{subfigure}
    \caption{Error analysis of false positives in VOC Novel Set 1 with and without decouple fine-tuning (DF). Detectors are trained with 3 shots. Pie charts indicate the fraction of correct detections (Cor) and top-ranked false positives that are due to poor localization (Loc), confusion with similar objects (Sim), confusion with other VOC objects (Oth), or confusion with background or unlabeled objects (BG).}
    \label{fig:df}
\end{figure}

\textbf{Decoupled fine-tuning reduces false positives.} We analyze the false positives generated by our SRR-FSD with and without decoupled fine-tuning (DF) using the detector diagnosing tool \cite{det-analysis}. The effect of DF on reducing the false positives in novel classes is visualized in Figure \ref{fig:df}. It shows that most of the false positives are due to misclassification into similar categories. With DF, the classification subnet can be trained independently from the localization subnet to learn better features specifically for classification.

\section{Conclusion}
In this work, we propose semantic relation reasoning for few-shot object detection. The key insight is to explicitly integrate semantic relation between base and novel classes with the available visual information, which can help to learn the novel concepts better especially when the novel class data is extremely limited. We apply the semantic relation reasoning to the standard two-stage Faster R-CNN and demonstrate robust few-shot performance against the variation of shot numbers. Compared to previous methods, our approach achieves state-of-the-art results on several few-shot detection settings, as well as a more realistic setting where novel concepts encoded in the pretrained backbone model are eliminated. 
We hope this realistic setting can be a better evaluation protocol for future few-shot detectors. Last but not least, the key components of our approach, i.e. semantic space projection and relation reasoning, can be straightly applied to the classification subnet of other few-shot detectors.

\appendix
\section{Removing Novel Classes from ImageNet}
\label{ap:real_setting}
We propose a realistic setting for evaluating the few-shot object detection methods, where novel classes are completely removed from the classification dataset used for training a model to initialize the backbone network in the detector. This can guarantee that the object concept of novel classes will not be encoded in the pretrained model before training the few-shot detector. Because the novel class data is so rare in the real world that pretraining a classifier on it is not realistic.

ImageNet \cite{imagenet} is widely used for pretraining the classification model. It has 1000 classes organized according to the WordNet hierarchy. Each class has over 1000 images for training. We systematically and hierarchically remove novel classes by finding each synset and its corresponding full hyponym (synset of the whole sub-tree starting from that synset) using the ImageNet API \footnote{http://image-net.org/download-API}. So each novel class may contain multiple ImageNet classes.

For the novel classes in the VOC dataset \cite{voc}, their corresponding WordNet IDs to be removed are as follows.
\begin{itemize}
    \item aeroplane: n02690373, n02692877, n04552348
    \item bird: n01514668, n01514859, n01518878, n01530575, n01531178, n01532829, n01534433, n01537544, n01558993, n01560419, n01580077, n01582220, n01592084, n01601694, n01608432, n01614925, n01616318, n01622779, n01795545, n01796340, n01797886, n01798484, n01806143, n01806567, n01807496, n01817953, n01818515, n01819313, n01820546, n01824575, n01828970, n01829413, n01833805, n01843065, n01843383, n01847000, n01855032, n01855672, n01860187, n02002556, n02002724, n02006656, n02007558, n02009229, n02009912, n02011460, n02012849, n02013706, n02017213, n02018207, n02018795, n02025239, n02027492, n02028035, n02033041, n02037110, n02051845, n02056570, n02058221
    \item boat: n02687172, n02951358, n03095699, n03344393, n03447447, n03662601, n03673027, n03873416, n03947888, n04147183, n04273569, n04347754, n04606251, n04612504
    \item bottle: n02823428, n03062245, n03937543, n03983396, n04522168, n04557648, n04560804, n04579145, n04591713
    \item bus: n03769881, n04065272, n04146614, n04487081
    \item cat: n02123045, n02123159, n02123394, n02123597, n02124075, n02125311, n02127052
    \item cow: n02403003, n02408429, n02410509
    \item horse: n02389026, n02391049
    \item motorbike: n03785016, n03791053
    \item sheep: n02412080, n02415577, n02417914, n02422106, n02422699, n02423022
    \item sofa: n04344873
\end{itemize}

For the novel classes in the COCO dataset \cite{coco}, they are very common in the real world. Removing them from the ImageNet does not make sense as much as removing data-scarce classes. So we suggest for large-scale datasets like COCO, we should follow the long-tail distribution of their class frequency and select the data-scarce classes on the distribution tail to be the novel classes.

\section{Visualization of Relation Reasoning}
\begin{figure}
    \centering
    \subfigure[Before relation reasoning]{
        \includegraphics[width=0.9\columnwidth]{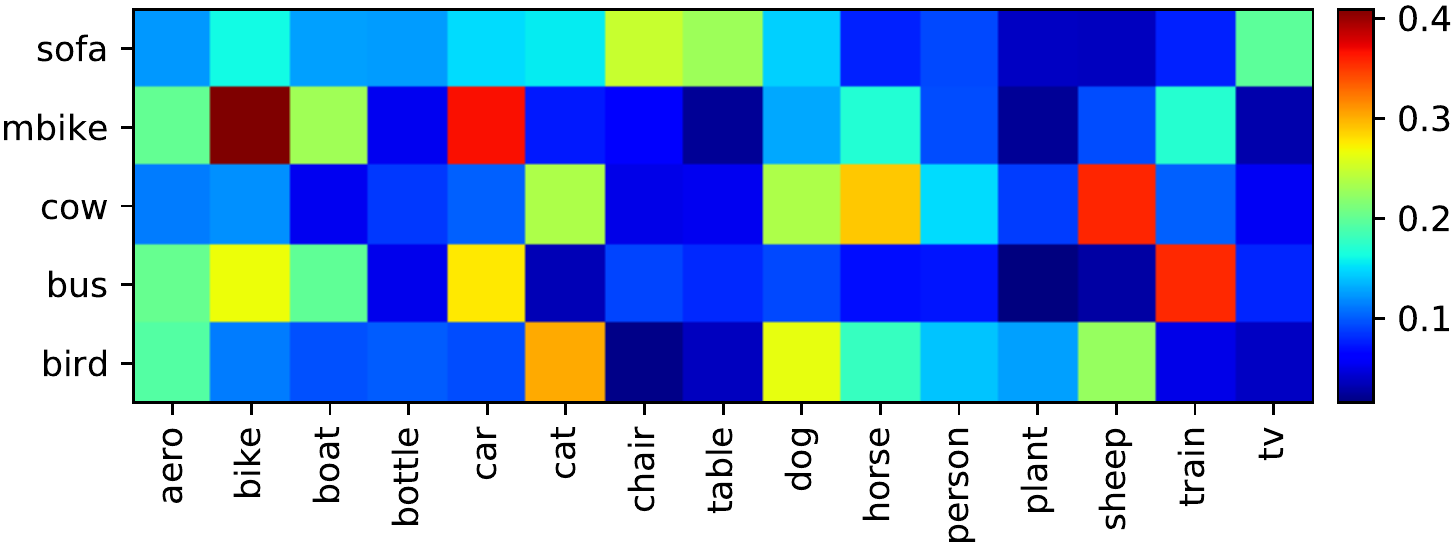}
        \label{fig:corr_sem}
    }
    \subfigure[After relation reasoning]{
        \includegraphics[width=0.9\columnwidth]{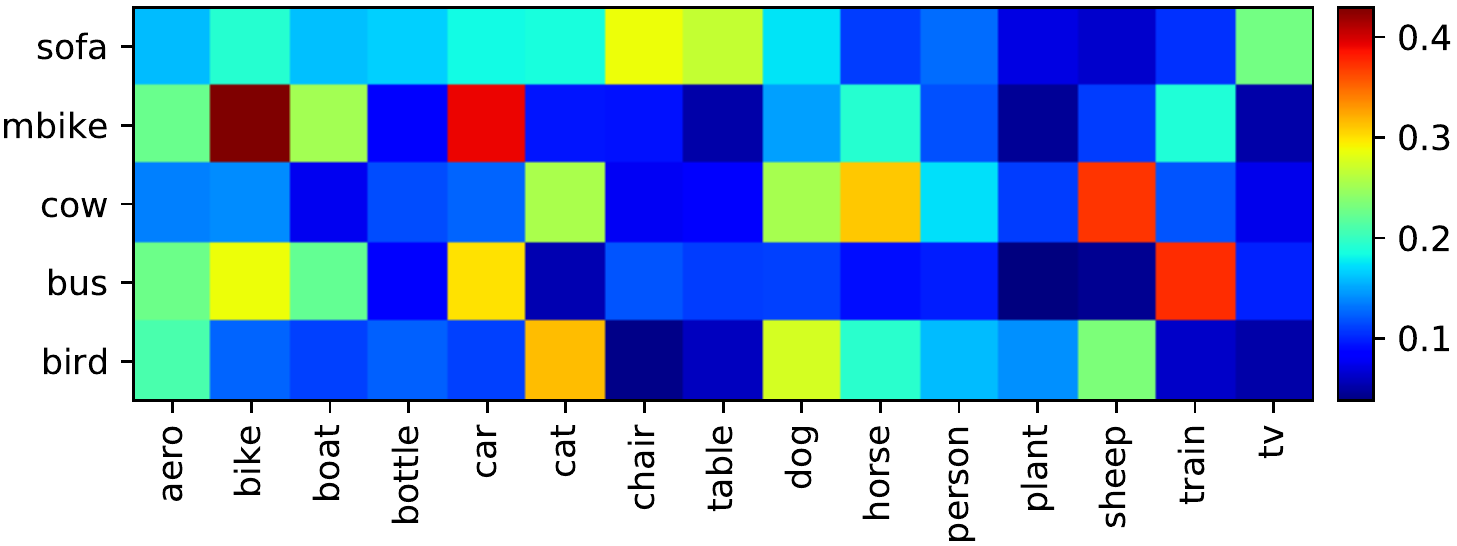}
        \label{fig:corr_rr}
    }
    \subfigure[Difference between above correlation maps]{
        \includegraphics[width=0.9\columnwidth]{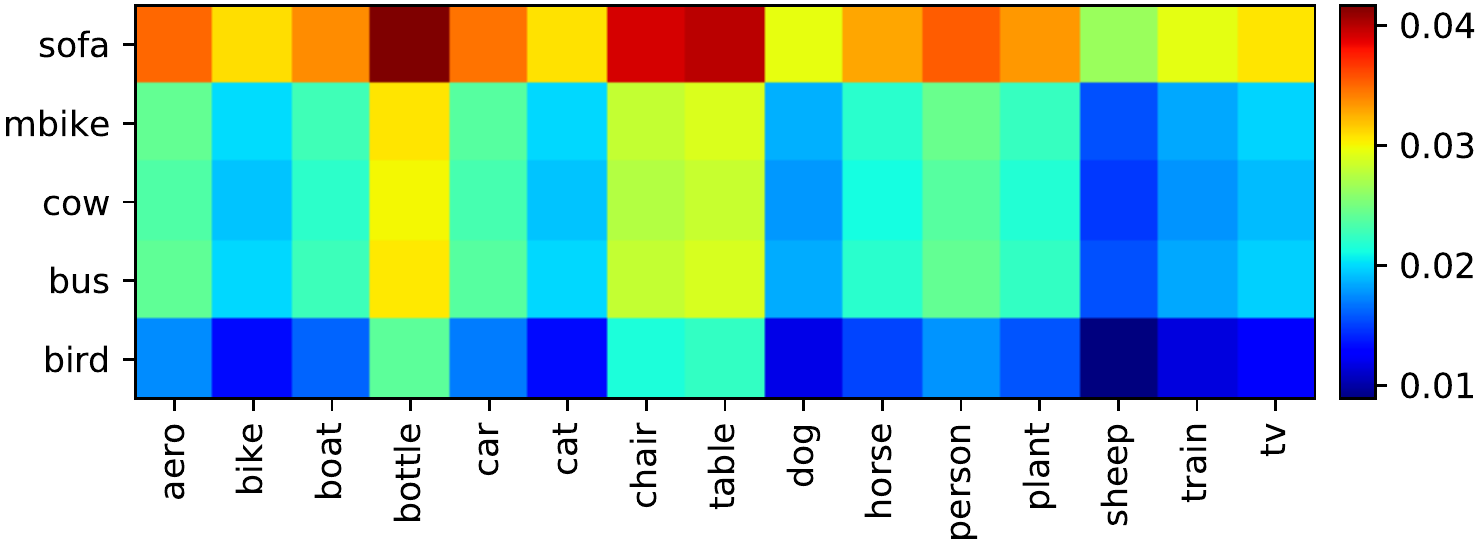}
        \label{fig:corr_delta}
    }
    \caption{Correlation of the semantic embeddings before and after the relation reasoning between the base classes and the novel classes on the VOC dataset. The novel classes are from Novel Set 1. The last figure shows how does the correlation change subtly. Some novel classes are getting more correlated with base classes after relation reasoning, e.g. ``sofa'' with ``bottle'' and ``table''. \textit{Best viewed in color.}
    }
    \label{fig:att_graph}
\end{figure}

Figure \ref{fig:att_graph} visualizes the correlation maps between the semantic embeddings of novel and base classes before and after the relation reasoning, as well as the difference between the two maps. Nearly all the correlations are increased slightly, indicating better knowledge propagation between the two groups of classes. Additionally, it is interesting to see that some novel classes get more correlated than others, e.g. ``sofa'' with ``bottle'' and ``sofa'' with ``table'', probably because ``sofa'' can often be seen together with ``bottle'' and ``table'' in the living room but the original semantic embeddings cannot capture these relationships.

\section{Using Other Word Embeddings}
In the semantic space projection, we represent the semantic space using word embeddings from the Word2Vec \cite{word2vec}. We could simply set the $\mathbf{W}_e$ to be random vectors. Additionally, there are other language models for obtaining vector representations for words, such as the GloVe \cite{glove}. The GloVe is trained with aggregated global word-word co-occurrence statistics from a corpus, and the resulting representations showcase interesting linear substructures of the word vector space. We also explored using word embedding with different dimensions from the GloVe in the semantic space projection step and compared with the results by the Word2Vec. Performance on the VOC Novel Set 1 is reported in Table \ref{tab:glove}. The Word2Vec can provide better representations than the GloVe of both 300 dimensions and 200 dimensions. The performance of random embeddings is significantly worse than the meaningful Word2Vec and GloVe, which again verifies the importance of semantic information for shot-stable FSOD.

\begin{table}
\setlength\tabcolsep{4pt}
\centering
\begin{tabular}{c|ccccc}
\hline \hline
                 & \multicolumn{5}{c}{Novel Set 1}  \\
Word embeddings & shot=1    & 2    & 3    & 5    & 10   \\ \hline
Random-300d & 33.2 & 37.5 & 43.0 & 47.0 & 51.5 \\
Word2Vec-300d \cite{word2vec}              & 42.8 & 47.1 & 49.0 & 50.8 & 52.8 \\
GloVe-300d \cite{glove}              & 38.8 & 44.8 & 46.6 & 49.0 & 54.3 \\
GloVe-200d \cite{glove}              & 39.7 & 44.6 & 45.8 & 49.4 & 53.0 \\ \hline
\end{tabular}
\caption{FSOD performance (mAP50) on the VOC Novel Set 1 under different word embeddings in the semantic space projection. All models are using the ResNet-50 network. 300d and 200d mean the numbers of embedding dimension are 300 and 200 respectively. The Word2Vec provides better representations than the GloVe.}
\label{tab:glove}
\end{table}

\section{Reduced Dimension in Relation Reasoning}
In the relation reasoning module, the dimension of word embeddings is reduced by linear layers before computing the attention map, which saves computational time. We empirically test different dimensions and select the one with the best performance, i.e. when the dimension is 32. But other choices are just slightly worse. Table \ref{tab:dimension} reports the results on VOC dataset under different dimensions. All the experiments are following the same setting as in the main paper. The only exception is that we use ResNet-50 \cite{resnet} to reduce the computational cost of tuning hyperparameters.

\begin{table}
\centering
\begin{tabular}{c|ccccc}
\hline \hline
                 & \multicolumn{5}{c}{Novel Set 1}  \\
Dimension & shot=1    & 2    & 3    & 5    & 10   \\ \hline
128              & 40.9 & 44.6 & 44.3 & 48.1 & 54.1 \\
64               & 42.0 & \textbf{47.4} & \textbf{48.9} & 51.7 & 54.1 \\
32               & \textbf{42.4} & \textbf{46.8} & \textbf{48.1} & \textbf{51.9} & \textbf{54.7} \\
16               & \textbf{44.1} & 46.0 & 47.8 & 51.7 & \textbf{54.7} \\ \hline
\end{tabular}
\caption{FSOD performance (mAP50) on the VOC Novel Set 1 under different reduced feature dimension in the relation reasoning module. Bold font indicates best or second best results. All models are using the ResNet-50 network. }
\label{tab:dimension}
\end{table}

\begin{table}
\setlength\tabcolsep{3pt}
\begin{tabular}{c|ccccc}
\hline \hline
                    & \multicolumn{5}{c}{Novel Set 1}   \\
Tunable Parameters  & shot=1 & 2    & 3    & 5    & 10   \\ \hline
Last layer (TFA \cite{tfa})    & 39.8   & 36.1 & 44.7 & 55.7 & 56.0 \\
+FCs                & 36.9   & 34.9 & 45.3 & 53.0 & 55.9 \\
+FCs  +RPN          & 37.2   & 39.8 & 44.3 & 52.7 & 56.2 \\
+FCs +RPN +Backbone & 16.2   & 19.5 & 24.8 & 39.2 & 44.6 \\ \hline
\end{tabular}
\caption{FSOD results (mAP50) on the VOC Novel Set 1 with more and more tunable parameters in the finetuning stage. The baseline is TFA \cite{tfa} which only finetunes the last classification layer in the Faster R-CNN. We gradually unfreeze more previous layers including two fully-connected layers (FCs) after the RoI-pooling, layers in region proposal network (RPN), and layers in the Backbone. This proves that finetuning more parameters does not guarantee better performance in few-shot detection.}
\label{tab:finetune}
\end{table}

\begin{table*}[t]
\centering
\setlength\tabcolsep{1.3pt}
\begin{tabular}{c|c|cccccc|cccccc|cccccc}
\hline \hline
\multicolumn{1}{c|}{}                   &                & \multicolumn{6}{c|}{Novel Set 1}                                                                                                                                & \multicolumn{6}{c|}{Novel Set 2}                                                                                                                                    & \multicolumn{6}{c}{Novel Set 3}                                                                                                                                 \\
\multicolumn{1}{c|}{Shot}               & Method         & \multicolumn{1}{c}{bird} & \multicolumn{1}{c}{bus} & \multicolumn{1}{c}{cow} & \multicolumn{1}{c}{mbike} & \multicolumn{1}{c}{sofa} & \multicolumn{1}{c|}{mean} & \multicolumn{1}{c}{aero} & \multicolumn{1}{c}{bottle} & \multicolumn{1}{c}{cow} & \multicolumn{1}{c}{horse} & \multicolumn{1}{c}{sofa} & \multicolumn{1}{c|}{mean} & \multicolumn{1}{c}{boat} & \multicolumn{1}{c}{cat} & \multicolumn{1}{c}{mbike} & \multicolumn{1}{c}{sheep} & \multicolumn{1}{c}{sofa} & \multicolumn{1}{c}{mean} \\ \hline
\multicolumn{1}{c|}{\multirow{4}{*}{1}} & FSRW           & 13.5 & 10.6 & 31.5 & 13.8 & 4.3 & 14.8 & 11.8 & \textbf{9.1} & 15.6 & 23.7 & 18.2 & 15.7 & 10.8 & 44.0 & 17.8 & 18.1 & 5.3 & 19.2 \\
 & Meta R-CNN     & 6.1 & 32.8 & 15.0 & 35.4& 0.2& 19.9& 23.9& 0.8& 23.6& 3.1& 0.7& 10.4& 0.6& 31.1& 28.9& 11.0& 0.1& 14.3 \\
 & MPSR           & 33.5& 41.2& 57.6& 54.5& 21.6& 41.7& 21.2& \textbf{9.1} & 36.0& 30.9& 25.1& 24.4& 14.9& 47.8& 57.7& 34.7& \textbf{22.8}& 35.6 \\
 & SRR-FSD (Ours) & \textbf{38.1} & \textbf{53.8} & \textbf{58.7} & \textbf{64.1} & \textbf{24.4} & \textbf{47.8} & \textbf{27.9} & 4.6 & \textbf{50.5} & \textbf{53.9} & \textbf{25.5} & \textbf{32.5} & \textbf{16.2} & \textbf{57.2} & \textbf{62.9} & \textbf{48.3} & 16.0 & \textbf{40.1} \\ \hline
\multirow{4}{*}{2} & FSRW           & 21.2& 12.0& 16.8& 17.9& 9.6& 15.5& 28.6& 0.9& 27.6& 0.0& 19.5& 15.3& 5.3& 46.4& 18.4& 26.1& 12.4& 21.7\\
 & Meta R-CNN     & 17.2& 34.4& 43.8& 31.8& 0.4& 25.5& 12.4& 0.1& 44.4& 50.1& 0.1& 19.4& 10.6& 24.0& 36.2& 19.2& 0.8& 18.2\\
 & MPSR           & \textbf{38.2}& 28.6& 56.5& 57.3& 32.0& 42.5& \textbf{36.5}& \textbf{9.1}& 45.1& 21.6& 34.2& 29.3& \textbf{17.9} & 49.6& 59.2& \textbf{49.2}& 32.9& \textbf{41.8}\\
 & SRR-FSD (Ours) & 35.8 & \textbf{57.7} & \textbf{59.3} & \textbf{61.8} & \textbf{38.0} & \textbf{50.5} & 34.4 & 5.7 & \textbf{57.1} & \textbf{44.0} & \textbf{35.5} & \textbf{35.3} & 15.5 & \textbf{51.4} & \textbf{62.6} & 44.4 & \textbf{33.7} & 41.5 \\ \hline
\multirow{4}{*}{3}      & FSRW           & 26.1& 19.1& 40.7& 20.4& 27.1& 26.7& 29.4& 4.6& 34.9& 6.8& 37.9& 22.7& 11.2& 39.8& 20.9& 23.7& 33.0& 25.7\\
 & Meta R-CNN     & 30.1& 44.6& 50.8& 38.8& 10.7& 35.0& 25.2& 0.1& 50.7& \textbf{53.2}& 18.8& 29.6& 16.3& 39.7& 32.6& 38.8& 10.3& 27.5 \\
 & MPSR           & 35.1& \textbf{60.6}& 56.6& 61.5& \textbf{43.4}& \textbf{51.4}& \textbf{49.2}& 9.1& 47.1& 46.3& \textbf{44.3}& \textbf{39.2}& 14.4& \textbf{60.6}& 57.1& 37.2& \textbf{42.3}& 42.3\\
 & SRR-FSD (Ours) & \textbf{35.2} & 55.6 & \textbf{61.3} & \textbf{62.9} & 41.5 & 51.3 & 42.3 & \textbf{11.5} & \textbf{57.0} & 43.6 & 41.2 & 39.1 & \textbf{23.1} & 50.6 & \textbf{60.0} & \textbf{49.3} & 38.6 & \textbf{44.3} \\ \hline
\multirow{4}{*}{5}      & FSRW           & 31.5 & 21.1 & 39.8 & 40.0 & 37.0 & 33.9 & 33.1 & 9.4 & 38.4 & 25.4 & 44.0 & 30.1 & 14.2 & 57.3 & 50.8 & 38.9 & 41.6 & 40.6\\
 & Meta R-CNN     & 35.8 & 47.9 & 54.9 & 55.8 & 34.0 & 45.7 & 28.5 & 0.3 & 50.4 & 56.7 & 38.0 & 34.8 & 16.6 & 45.8 & 53.9 & 41.5 & 48.1 & 41.2\\
 & MPSR           & 39.7 & \textbf{65.5} & 55.1 & \textbf{68.5} & \textbf{47.4} & 55.2 & \textbf{47.8} & 10.4 & 45.2 & 47.5 & \textbf{48.8} & 39.9 & \textbf{20.9} & \textbf{56.6} & \textbf{68.1} & 48.4 & \textbf{45.8} & \textbf{48.0}\\
 & SRR-FSD (Ours) & \textbf{46.1} & 58.6 & \textbf{64.6} & 63.5 & 43.2 & \textbf{55.2} & 44.2 & \textbf{12.3} & \textbf{56.5} & 51.3 & 39.8 & \textbf{40.8} & 20.4 & 55.5 & 65.4 & \textbf{51.9} & 41.3 & 46.9 \\ \hline
\multirow{4}{*}{10} & FSRW           & 30.0 & 62.7 & 43.2 & 60.6 & 39.6 & 47.2 & 43.2 & 13.9 & 41.5 & 58.1 & 39.2 & 39.2 & 20.1 & 51.8 & 55.6 & 42.4 & 36.6 & 41.3 \\
 & Meta R-CNN     & \textbf{52.5} & 55.9 & 52.7 & 54.6 & 41.6 & 51.5 & \textbf{52.8} & 3.0 & 52.1 & \textbf{70.0} & 49.2 & 45.4 & 13.9 & \textbf{72.6} & 58.3 & 47.8 & 47.6 & 48.1\\
 & MPSR           & 48.3 & \textbf{73.7} & \textbf{68.2} & \textbf{70.8} & \textbf{48.2} & \textbf{61.8} & 51.8 & 16.7 & 53.1 & 66.4 & \textbf{51.2} & \textbf{47.8} & \textbf{24.4} & 55.8 & \textbf{67.5} & \textbf{50.4} & \textbf{50.5} & \textbf{49.7} \\
 & SRR-FSD (Ours) & 45.0 &	67.4 &	63.1 &	65.2 &	43.3 &	56.8 & 46.2 &	\textbf{18.4} &	\textbf{54.0} &	59.1 &	41.4 &	43.8 & 17.1 &	55.1 &	67.4 &	47.5 &	44.7 &	46.4\\ \hline
\end{tabular}
\caption{AP50 performance of each novel class on the few-shot VOC dataset. Bold font indicates the best result in the group. Our SRR-FSD trained with visual information and semantic relation demonstrates shot-stable performance.}
\label{tab:voc_complete}
\end{table*}

\section{Finetuning More Parameters}
\label{ap:finetune}
Similar to TFA \cite{tfa}, we have a finetuning stage to make the detector generalized to novel classes. For the classification subnet, we finetune the parameters in the relation reasoning module and the projection matrix while all the parameters in previous layers are frozen. Some may argue that the improvement of our SRR-FSD over the baseline is due to more parameters finetuned in the relation reasoning module compared to the Faster R-CNN \cite{faster-rcnn} baseline. But we show that finetuning more parameters does not necessarily lead to better results in Table \ref{tab:finetune}. We take the TFA model which is essentially a Faster R-CNN finetuned with only the last layer trainable and gradually unfreeze the previous layers. It turns out more parameters involved in finetuning do not change the results substantially and that too many parameters will lead to severe overfitting.

\section{Complete Results on VOC}
In Table \ref{tab:voc_complete}, we present the complete results on the VOC \cite{voc} dataset as in FSRW \cite{yolo-fewshot} and Meta R-CNN \cite{meta-rcnn}. We also include the very recent MPSR \cite{mpsr} for comparison. MPSR develops an auxiliary branch to generate multi-scale positive samples as object pyramids and to refine the prediction at various scales. Note that MPSR improves its baseline by a considerable margin but its research direction is \textit{orthogonal and complimentary} to ours because it is still exclusively dependent on visual information. Therefore, our approach combining visual information and semantic relation reasoning can achieve superior performance at extremely low shot (e.g. 1, 2) conditions.

\section{Interpretation of the Dynamic Relation Graph}
In the relation reasoning module, we propose to learn a \textit{dynamic} relation graph driven by the data, which is conceptually different from the predefined fixed knowledge graphs used in \cite{zsr-gnn, multi-label-gcn, fsr-kt}. We implement the dynamic graph with the self-attention architecture \cite{attention}. Although it is in the form of a feedforward network, it can also be interpreted as a computation related to the knowledge graph. If we denote the transformations in the linear layers $f$, $g$, $h$, $l$ as $\mathbf{T}_f$, $\mathbf{T}_g$, $\mathbf{T}_h$, $\mathbf{T}_l$ respectively, we can formulate the relation reasoning in Eq. \eqref{eq:attention}
\begin{equation}
    \mathbf{W}_e' = \delta(\mathbf{W}_e \mathbf{T}_f \mathbf{T}_g^T \mathbf{W}_e^T) \mathbf{W}_e \mathbf{T}_h \mathbf{T}_l + \mathbf{W}_e
    \label{eq:attention}
\end{equation}
where $\mathbf{W}_e'$ is the matrix of augmented word embeddings after the relation reasoning which will be used as the weights to compute classification scores and $\delta$ is the softmax function operated on the last dimension of the input matrix. The item $\delta(\mathbf{W}_e \mathbf{T}_f \mathbf{T}_g^T \mathbf{W}_e^T)$ can be interpreted as a $N \times N$ dynamic knowledge graph in which the learnable parameters are $\mathbf{T}_f$ and $\mathbf{T}_g$. And it is involved in the computation of the classification scores via the graph convolution operation \cite{gcn}, which connects the $N$ word embeddings in $\mathbf{W}_e$ to allow knowledge propagation among them. The item $\mathbf{T}_h \mathbf{T}_l$ can be viewed as a learnable transformation applied to each embedding independently.

{\small
\bibliographystyle{ieee_fullname}
\bibliography{egbib}
}

\end{document}